\documentclass[conference,10pt]{IEEEtran}
\IEEEoverridecommandlockouts
\AtBeginDocument{%
  }

\usepackage{booktabs}
\usepackage{multirow}
\usepackage{hyperref}
\usepackage{comment}
\usepackage{xcolor}
\usepackage{adjustbox}
\usepackage{amsmath}
\usepackage{amsfonts} 
\usepackage{threeparttable}

\newcommand{\shortsection}[1]{\vspace*{1ex}\noindent{\bf #1:}}
\begin{document}
\bstctlcite{IEEEexample:BSTcontrol}

\title{Real-Time Multimodal Cognitive Assistant for \\ Emergency Medical Services}

\makeatletter
\newcommand{\linebreakand}{%
  \end{@IEEEauthorhalign}
  \hfill\mbox{}\par
  \mbox{}\hfill\begin{@IEEEauthorhalign}
}
\makeatother

\author{\IEEEauthorblockN{Keshara Weerasinghe}
\IEEEauthorblockA{\textit{Electrical and Computer Engineering} \\
\textit{University of Virginia}\\
cjh9fw@virginia.edu
}
\and
\IEEEauthorblockN{Saahith Janapati}
\IEEEauthorblockA{\textit{Computer Science} \\
\textit{University of Virginia}\\
jax4zk@virginia.edu
% email address or ORCID
}
\and
\IEEEauthorblockN{Xueren Ge}
\IEEEauthorblockA{\textit{Electrical and Computer Engineering} \\
\textit{University of Virginia}\\
zar8jw@virginia.edu
% email address or ORCID
}
\linebreakand
\IEEEauthorblockN{Sion Kim}
\IEEEauthorblockA{\textit{Computer Science} \\
\textit{University of Virginia}\\
sk9uth@virginia.edu
% email address or ORCID
}
\and
\IEEEauthorblockN{Sneha Iyer}
\IEEEauthorblockA{\textit{Computer Science} \\
\textit{University of Virginia}\\
ssi3ka@virginia.edu
% email address or ORCID
}

\and
\IEEEauthorblockN{John A. Stankovic}
\IEEEauthorblockA{\textit{Computer Science} \\
\textit{University of Virginia}\\
jas9f@virginia.edu
% email address or ORCID
}
\and
\IEEEauthorblockN{Homa Alemzadeh}
\IEEEauthorblockA{\textit{Electrical and Computer Engineering} \\
\textit{University of Virginia}\\
ha4d@virginia.edu
% email address or ORCID
}
}

%   \includegraphics[width=\textwidth]{figures/iotdi-teaser.pdf}
%   \caption{First Responders}
%   \Description{Doing something REPLACE}


\maketitle
\thispagestyle{plain}
\pagestyle{plain}

\begin{abstract}

%Emergency Medical Services (EMS) serve as the primary response for delivering swift medical assistance to those in need. 
Emergency Medical Services (EMS) responders often operate under time-sensitive conditions, facing cognitive overload and inherent risks, requiring essential skills in critical thinking and rapid decision-making. This paper presents CognitiveEMS, an end-to-end wearable cognitive assistant system that can act as a collaborative virtual partner engaging in the \textit{real-time} acquisition and analysis of \textit{multimodal data} from an emergency scene and interacting with EMS responders through Augmented Reality (AR) smart glasses. CognitiveEMS processes the continuous streams of data in real-time and leverages edge computing to provide assistance in EMS protocol selection and intervention recognition. We address key technical challenges in real-time cognitive assistance by introducing three novel components:
(i) a Speech Recognition model that is fine-tuned for real-world medical emergency conversations using simulated EMS audio recordings, augmented with synthetic data generated by large language models (LLMs); (ii) an EMS Protocol Prediction model that combines state-of-the-art (SOTA) tiny language models with EMS domain knowledge using graph-based attention mechanisms; (iii) an EMS Action Recognition module which leverages multimodal audio and video data and protocol predictions to infer the intervention/treatment actions taken by the responders at the incident scene. Our results show that for speech recognition we achieve superior performance compared to SOTA (WER of \textbf{0.290} vs. \textbf{0.618}) on conversational data. Our protocol prediction component also significantly outperforms SOTA (top-3 accuracy of \textbf{0.800} vs. \textbf{0.200}) and the action recognition achieves an accuracy of \textbf{0.727}, while maintaining an end-to-end latency of \textbf{3.78s} for protocol prediction on the edge and \textbf{0.31s} on the server.

\end{abstract}

\section{Introduction}
Emergency Medical Services (EMS) responders provide life support and stabilization for patients before transporting them to a hospital. They first decide on the appropriate emergency response protocol to follow based on their assessment at the incident scene, then execute the medical interventions (treatments such as Cardiopulmonary Resuscitation (CPR) and medications) according to the protocol. They often make critical decisions under time-sensitive conditions and cognitive overload~\cite{cognitiveload,preum2018towards,preum2019cognitiveems}. The right selection of the emergency response protocol is crucial but demanding, requiring the processing of information from many different sources under pressure. Even after selecting the right protocol, responders still need to be on high alert as they execute interventions and monitor the patient's response. Finally, they need to recall all details of the incident and report them in a post-incident form.

Several previous works have focused on developing assistive technologies and cognitive assistant systems to help first responders with decision making~\cite{preum2019cognitiveems,shu2019behavior,emsreact,jin2023emsassist} and information collection~\cite{emscontext, rahman2020grace,kim2021information}. However, there are still several challenges and research gaps to be addressed. 

% - Reliance on Behavior Trees and State Machines
\textbf{Challenge of Modeling Domain Knowledge:} Previous works have developed pre-defined rule-based systems~\cite{preum2019cognitiveems}, Behavior Trees (BTs)~\cite{shu2019behavior}  and state machines~\cite{emsreact} for modeling the knowledge on a limited set of EMS protocols and interventions for decision support. These systems, although interpretable, are not easily scalable because of the large amount of time and expert knowledge needed to manually encode and update their models. On the other hand, large deep learning and language models could provide generalization and scalability at the risk of losing transparency \cite{shu2019behavior}. %In addition, they could easily become complex with thousands of rules and states, making it difficult to understand and maintain. 
There exists a need for scalable and explainable models that capture domain knowledge while supporting a diverse range of scenarios.

% - Limited Protocol prediction
%\cite{emsreact} was developed specifically for training purposes and a specific protocol only. There exists a need for a system that can assist responders through the diverse range of incidents they encounter and the respective protocols. 

% - Not deployed on edge 
\textbf{Challenge of Unreliable Communication:} Existing cognitive assistant systems leverage cloud-based models and services for transcribing audio from responder conversations into text (e.g., Google Speech-to-Text)~\cite{preum2018towards,preum2019cognitiveems} and medical information extraction and protocol inference (e.g., MetaMap)~\cite{emscontext, preum2019cognitiveems}. However, responders are often called to areas or situations with unreliable network connections, such as rural areas, sites of catastrophe, and natural disasters. A reliance on the cloud is a liability for a system designed to operate in such scenarios.

% unimodal system
\textbf{Challenge of Noisy and Incomplete Sensor Data:} Previous works have mainly focused on developing unimodal systems based on audio data and speech recognition for assisting responders. However, audio in emergency scenes is prone to environmental noise that severely affects the quality of data and accuracy of speech recognition~\cite{preum2019cognitiveems,rahman2020grace,emsreact}. Also, audio data from responder conversations might not capture all the critical events and observations from the incident scene. %While there is nothing inherently at fault with unimodal systems, multimodal systems allow for the collection of new information that was unavailable previously in an unimodal system. 
Fusing other modalities such as vision and hand activity data~\cite{rahman2023senseems} could enable a more holistic understanding of the incident scene and more insightful feedback to responders.

% Not real-time and continous
\textbf{Challenge of Real-time Cognitive Assistance at the Edge:} Existing EMS cognitive assistants, such as EMSAssist~\cite{jin2023emsassist}, rely on one-time audio recordings as input to provide one-time protocol predictions as output. But in reality, incidents progress quickly with dynamic levels of information to be processed and decisions to be made. Therefore, reliance on recorded audio as one-time input is insufficient for supporting responders' continuous decision-making in real-time. 
% - Not real-time and continuous
In addition, previous works utilize unrealistic non-conversational speech data to evaluate their systems~\cite{jin2023emsassist}. However, responders' conversations are verbalized full-phrased observations and essential information about the scene and patient status~\cite{shu2019behavior}.

% CongitiveEMS Contribution
To address these challenges, we present CognitiveEMS, a \textit{real-time multimodal} cognitive assistant system that provides \textit{continuous end-to-end} support to responders through protocol prediction and intervention recognition at the \textit{edge}.

Our contributions are the following: 
\begin{itemize}
    \item Real-time hands-free cognitive assistance at the edge through wireless streaming and parallel processing of multimodal data (audio and vision) from smart glasses.
    % \item New publicly available EMS audio dataset consisting of Human and Synthetic conversational Audio and a novel method for generating synthetic medical conversational audio leveraging LLMs.   
    %\item A Novel method for generating synthetic EMS conversational audio leveraging LLMs and domain knowledge.
    \item EMS-Whisper, a speech recognition model fine-tuned on realistic EMS audio recordings and synthetic EMS conversational audio, generated using a text-to-speech model and an LLM prompted with domain knowledge, for improved real-time transcription at the edge (with WER of \textbf{0.290} vs. SOTA's \textbf{0.618}).
    \item EMS-TinyBERT, a tiny language model integrated with EMS domain knowledge and fine-tuned with real EMS data for protocol prediction (with a significantly higher top-3 accuracy of \textbf{0.800} vs. \textbf{0.200} than SOTA).
    \item Real-time EMS intervention recognition at the edge using protocol knowledge and zero-shot image classification (with end-to-end accuracy of \textbf{0.727}).
    \item End-to-end performance and latency evaluation of the overall system both on an edge device and a server with multimodal data from simulated EMS scenarios. 
    \item New publicly-available datasets, including human and synthetic EMS conversational audio, multimodal data from simulated EMS scenarios, and an open-source codebase, online at \url{https://github.com/UVA-DSA/EMS-Pipeline}.
    % \href{https://github.com/UVA-DSA/CognitiveEMS.git}{GitHub}. 
\end{itemize}

\section{Background and Related Work}

%\subsection{EMS Protocols and Interventions}
%In an EMS incident, responders follow a certain protocol, executing a specific set of steps and actions called interventions based on the patient's condition and status. Upon arrival, responders interact with the patient and bystanders to collect more information about the incident. Then, based on all the information they collected, responders choose the appropriate EMS protocol and follow its interventions to stabilize the patient and transport them to hospital care. 
Upon arrival at an incident scene, EMS responders interact with the patient and bystanders to collect information about the incident. Then, based on the patient's condition and status they choose an appropriate set of EMS protocols to follow. Each EMS protocol defines a specific set of steps and actions called interventions to stabilize the patient before being transported to the hospital. 
Various EMS agencies adhere to protocol guidelines established by regional planning agencies. In this work, we follow the protocol guidelines delineated by the Old Dominion EMS Alliance (ODEMSA)\footnote{\url{https://odemsa.net/}}. 

Deciding on the correct protocol is essential for the health and well-being of the patient and is arguably the most important and effortful task taken by the responders that requires experience and domain knowledge. Imagine a scenario where responders arrive at a shopping mall to find a middle-aged patient lying on the floor surrounded by concerned bystanders. Bystanders testify that the patient suddenly collapsed after complaining of chest pain. Upon quick examination, responders find the patient is unresponsive to painful stimuli and without a pulse or respiration. Based on this information, the responders choose the Cardiac Arrest Protocol~\cite{ODEMSACardiac} and conduct the proper interventions in a specific order and duration: start CPR, check rhythm, administer drugs, and support the airway (as illustrated in Figure \ref{fig:Cardiac Arrest Protocol}). Like the Cardiac Arrest protocol, all EMS protocols are dynamic in nature. Responders need to continuously make decisions and follow different intervention paths based on how the patient's condition evolves. %The dynamic nature of protocols calls for real-time system that can continuously takes in inputs over time, allowing the system to be aware of changes in incident circumstances rather than a static system that finishes running with one-time input\cite{jin2023emsassist}. To our knowledge, we implement the first Cognitive Assistant system in the EMS domain that is real-time, i.e., a system that continuously takes in input at a constant frequency and outputs timely protocol suggestions.

%So developing a system for real-time identification of protocols and interventions and providing feedback to responders at the incident scene requires continuous tracking of the incident circumstances, including patient signs and symptoms and responders' actions, through sensor measurements.

Responders document the EMS incidents in the form of electronic Patient Care Reports, referred to as ePCR. ePCR includes information similar to a patient's Electronic Health Records (EHR), such as a textual narrative describing the observations made and actions taken by the responders during the incident, as well as structured data on call type, chief complaints, impressions, procedures, and medications~\cite{kim2021information,shu2019behavior}. 

\begin{figure}[b!] % Use the figure* environment for figures that span two columns
    \vspace{-1em}
    \includegraphics[width=0.44\textwidth]{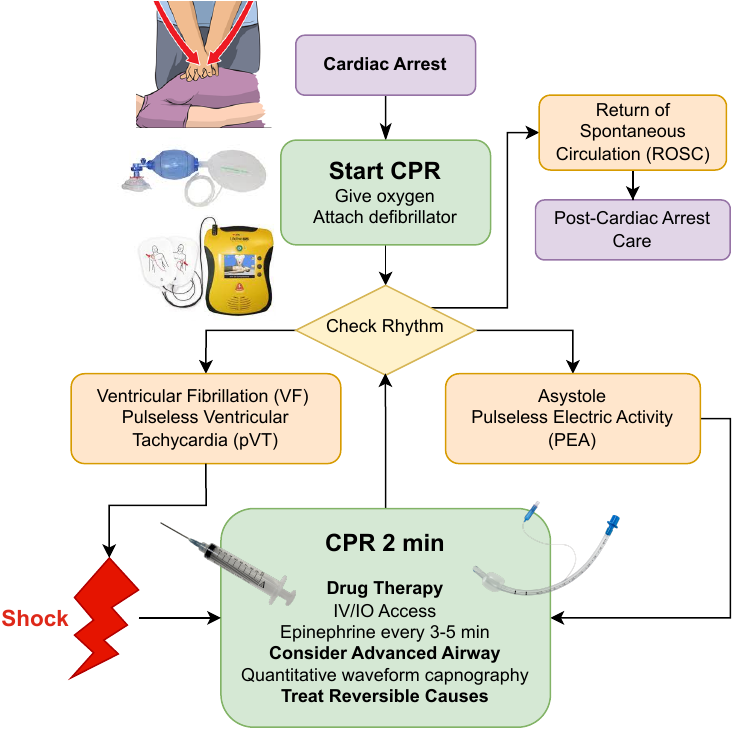}
    \caption{Responders Continuously Make Decisions During the Cardiac Arrest Protocol (Based on ODEMSA Document~\cite{ODEMSACardiac})}
    \label{fig:Cardiac Arrest Protocol}
\end{figure}

\subsection{EMS Cognitive Assistants}

Cognitive assistants are context-aware and adaptive systems that augment the cognitive capabilities of their users~\cite{preum2021review}. Previous works have introduced unimodal EMS cognitive assistant systems that process speech from an incident scene to provide first responders with suggestions and feedback regarding the protocols and associated interventions they perform at a scene~\cite{shu2019behavior, jin2023emsassist} or to help the responders with automated information extraction and filling of ePCR~\cite{emscontext,rahman2020grace}. These cognitive assistants are limited in their ability to accurately capture the incident context and provide meaningful feedback to responders because speech is not sufficient to obtain a holistic understanding of an emergency scenario. Crucial information is often not explicitly verbalized by responders and patients; instead, it is communicated through other modalities, such as visual signals and cues. For example, the presence or use of specific objects and devices (e.g., ECG monitors, AED devices, etc.) at the incident scene or treatment actions performed by the responders (e.g., clinical procedures such as CPR, Oxygen Administration, Epinephrine Administration) might not be verbalized by the responders. %As a result, a cognitive assistant capable of integrating multiple modalities into its understanding of an emergency scene can provide more insightful and relevant support to first responders.
Analysis of multimodal data (e.g., speech and video) from an incident scene can enable the recognition of important objects and actions and more accurate detection of protocols and interventions, allowing timely relevant feedback for improving responders' skills and performance in critical situations. This could be helpful in training where detailed and explainable feedback on executing interventions according to the protocols is needed.

%{The cognitive assistant system enhances its situational awareness by analyzing multimodal data, enabling it to provide informed feedback to emergency responders. The system's intervention recognition module plays a crucial role by integrating visual analysis with protocol predictions and prior knowledge, allowing for timely and relevant feedback that improves responder performance in critical situations.}
%Our multimodal cognitive assistant, CognitiveEMS, allows us to leverage multiple data sources to assist emergency responders on a level that has not been achieved before. Our CognitiveEMS pipeline can recognize interventions utilizing protocol knowledge and multimodal data input, extending cognitive assistance to a level beyond the current SOTA. 
To our knowledge, CognitiveEMS is the first real-time multimodal cognitive assistant system in the EMS domain which leverages both audio and video data from the incident scene for protocol prediction and subsequent intervention recognition.

\begin{figure*}[] % Use the figure* environment for figures that span two columns
    \centering
    \includegraphics[width=\textwidth]{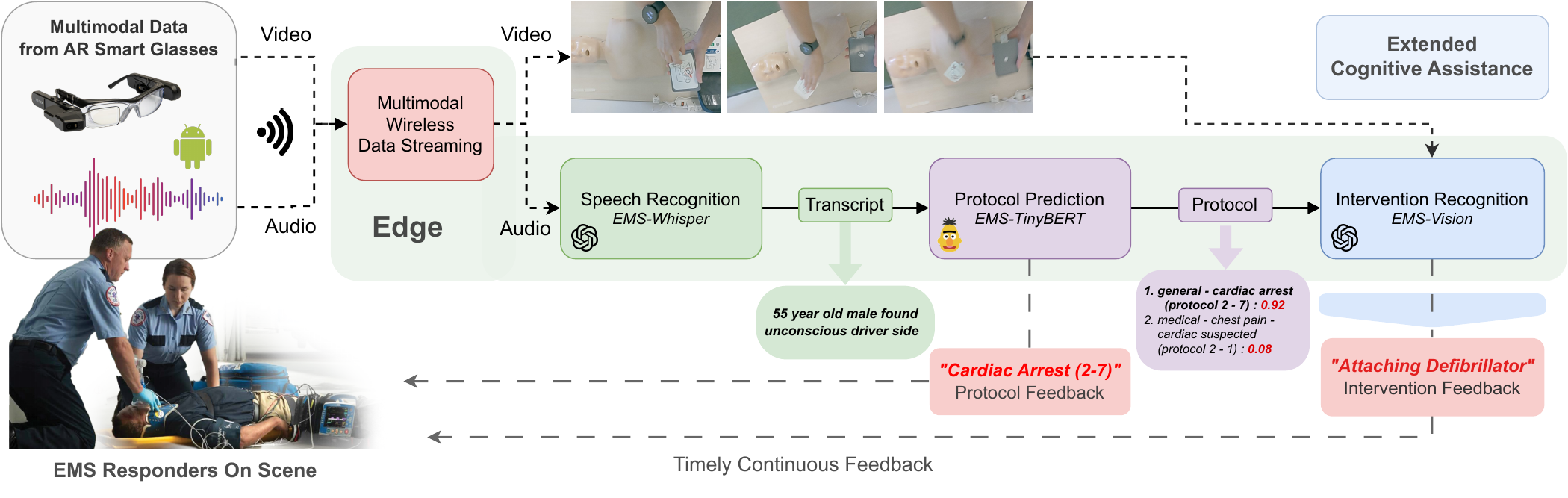} % Replace 'example-image' with your image file name and provide the correct path.
    \vspace{-2em}
    \caption{Overall Architecture of the Real-Time Multimodal CognitiveEMS Pipeline}
    \label{fig:overall_arch}
    \vspace{-0.5em}
\end{figure*}

\subsection{Automatic Speech Recognition}
Modern day Automatic Speech Recognition (ASR) models leverage deep learning architectures such as the Transformer~\cite{vaswani2017attention} and are trained using supervised and unsupervised learning techniques. Conformer~\cite{gulati2020conformer} and Whisper~\cite{radford2023robust} are both models that utilize the Transformer architecture. In the domain of transcription for medical audio,~\cite{47001} has developed specialized ASR models solely trained on speech from the medical domain. This requires the curation of a large medical-focused dataset, which is difficult due to financial constraints and patient privacy restrictions. A more feasible approach found to yield improved transcription performance is to fine-tune an existing pretrained model using a (relatively) small amount of domain-specific audio~\cite{luo2021cross}. In addition, synthetic audio has previously been shown to improve the performance of ASR models in specific domains. For example, SynthASR proposes using synthetically generated audio to adapt existing ASR models for specific applications~\cite{fazel2021synthasr}. 
% \cite{zheng2021using} demonstrates that fine-tuning ASR models with synthetically generated audio of out-of-vocabulary (OOV) words can boost performance of ASR systems on such words \cite{zheng2021using}. 

Within the EMS domain,~\cite{preum2019cognitiveems} utilizes cloud services (e.g., Google Cloud Speech-To-Text) for automatic speech recognition. While cloud services offer improved performance via larger models, they are a liability in a cognitive assistant system because they rely on stable network connections, which can be weak or nonexistent where responders are called.~\cite{jin2023emsassist} proposes the EMSConformer, a fine-tuned Conformer model for offline ASR model for EMS. We find several limitations in this work. First, the mobile application developed in~\cite{jin2023emsassist} requires responders to manually record and upload their speech on the mobile application, which may increase the cognitive overload on responders. Second, incidents progress at a dynamic and quick rate, with new information to process arising continuously from start to end. Due to the reliance on one-time recorded audio as input, EMSConformer is incapable of transcribing constantly changing information and supporting the decision-making of responders in real-time. Third, EMSConformer has been evaluated on unrealistic and non-conversational speech input. The evaluation data consists of isolated medical terminology and fragmented groups of words. However, responders' conversations and verbalized observations during real incidents will be fully-formed phrases and include words outside of medical terminology to describe signs and symptoms, such as saying, ``The patient says they feel a ton of bricks on their chest" instead of saying ``chest pain"~\cite{kim2021information}. Good-quality transcription of conversational audio typically heard at EMS scenes is crucial to obtain a holistic understanding of the incident scene and the status of patients. To our knowledge, EMS-Whisper is the first offline ASR model for EMS conversational data that can continuously transcribe an audio stream, allowing for effective edge deployment of an EMS cognitive assistant. In addition, we develop a novel method for generating synthetic EMS conversational audio to fine-tune pretrained ASR models.

\subsection{Multi-Label Text Classification}
Automatic medical diagnosis is the task of assigning diagnosis codes to a patient record based on free-form text. 
The task of assigning the most relevant subsets of labels to an instance is known as multi-label text classification (MLTC) in NLP. There is extensive prior research on MLTC in the medical domain, mainly for ICD code classification (a medical classification list designated by the World Health Organization) based on EHR data.~\cite{rios2018few, mullenbach2018explainable} utilized graph neural networks to combine domain knowledge with BERT models~\cite{devlin2018BERT} to find the most informative n-grams in the EHR for each ICD9 code. However, most prior works developed models for the clinical domain and cannot be directly applied to the EMS protocol prediction for several reasons. First, there is a domain mismatch between EHR and EMS data (ePCR). Compared with EHR, ePCR and EMS transcripts are collected over a short time frame and contain unconfirmed diagnoses, domain-specific terminology, and abbreviations. Second, most previous works for clinical ICD code classification ignore the impact of model size and inference latency when deployed in real-world scenarios on edge devices. For protocol prediction in EMS domain,~\cite{shu2019behavior} proposed a weakly supervised approach to determine the relevance between the current situation at the scene and each EMS protocol by calculating the similarity between their feature vectors. \cite{ge2023dkec} proposed to extract medical entities from EMS domain knowledge to build a heterogeneous graph and fuse domain knowledge with text features to compensate for EMS data scarcity problem. In~\cite{jin2023emsassist}, a deep learning based approach, called EMSMobileBERT, was proposed by fine-tuning light-weighted MobileBERT(25M) on an EMS corpus. However, the input text to EMSMobileBERT is in the form of symptom-based medical phrases which is not representative of real EMS scenarios where textual narratives from ePCR or transcribed conversational audio are expected. To our knowledge, our proposed model, EMS-TinyBERT, is the first real-time protocol prediction model that can process realistic textual transcripts (from ePCR or transcribed audio) as input, with a significantly higher top-3 accuracy than SOTA. 

% \colorbox{yellow}{Should the following text be here (in the background section)?}We take a similar approach to \cite{ge2023dkec} by integrating EMS domain knowledge to a fine-tuned a tiny language model which is pre-trained on medical corpus. We then deploy this model on our edge device.

\subsection{Human Action Recognition}
Human Action Recognition (HAR) has been extensively studied, with applications in numerous domains such as video surveillance, entertainment, sports,~\cite{kong2022human} and medicine~\cite{schrader2020advanced,kantoch2017human,mukherjee2020ensemconvnet}. Previous work has explored various sources of data,~\cite{pareek2021survey} including RGB images, depth information, and wearable sensors. However, within the EMS domain, there is limited work on responder action and intervention recognition.~\cite{zhang2023human} proposes a method that utilizes a spatial-temporal fusion convolution neural network (CNN) to recognize actions taken place during emergency incidents by responders. However, like most previous works in other domains, this method relies on a significant volume of annotated EMS data~\cite{wang2019survey}, which is notably scarce and costly to obtain. Recently, zero-shot learning methods have overcome traditional supervised learning challenges, including the need for extensive annotations~\cite{wang2019survey} and the inability to generalize to different classes~\cite{xian2017zero}. %Zero-shot models are gaining popularity in domains like computer vision and natural language processing. 
For instance, CLIP~\cite{radford2021learning} excels in zero-shot image classification by associating images with relevant natural language captions. It is trained by contrastive learning based on a large dataset of image-text pairs from the Internet. 
%To our knowledge, we are the first utilize protocol knowledge and zero-shot image classification models to recognize EMS interventions real-time, extending cognitive assistance for responders. Intervention recognition provides many benefits and applications: logging and form-filling, training EMS responders with real-time feedback, and verification of protocol procedure executions during actual incidents. Responders may make mistakes during interventions due to cognitive overload, and verification of interventions in real-time can assist and prevent responders from doing so. Another source of cognitive load for responders is filling out post-incident forms, and having a cognitive assistant track and log interventions can streamline this tedious process.

This paper presents our preliminary work on real-time recognition of EMS interventions using the knowledge of protocol guidelines and zero-short image classification. The real-time recognition of responders' actions enables monitoring of the critical aspects of interventions (e.g., CPR rate and depth \cite{rahman2023senseems}) and validating them for adherence to established protocol guidelines during training or actual incidents to enhance responders' skills and quality of care. It can also assist in the automated logging of EMS data and ePCR filing \cite{rahman2020grace}. %To our knowledge, we are the first to utilize protocol knowledge and zero-shot image classification to recognize EMS interventions in real-time.  

\section{Real-time Cognitive Assistant Pipeline}
This section presents the overall architecture of the CognitiveEMS pipeline as illustrated in Figure \ref{fig:overall_arch}.
%\subsection{Real-time Cognitive Assistant Pipeline}
Our primary objective is the creation of a system designed to provide uninterrupted, real-time assistance to EMS responders. As depicted in Figure \ref{fig:rt_pipeline}, we develop the CognitiveEMS as a real-time, multi-threaded architecture, aiming for a Service Level Objective (SLO) of a 4-second response interval for delivering protocol prediction feedback. This design criterion is pivotal for offering prompt support to responders amidst critical situations, with the system delivering continuous feedback at 4-second intervals based on the most recent data inputs. The establishment of this SLO was informed by consultations with our Emergency Medical Technician (EMT) collaborators \cite{rahman2023emsreact} and by considering the response times for critical medical emergencies and current hardware constraints. %Although a 3-second SLO would be ideal for our cognitive assistant to ensure optimal responsiveness, current hardware constraints necessitate a 4-second SLO. 
Future work will focus on overcoming current limitations to achieve better response times and further improve the effectiveness and timeliness of feedback to responders.

Response time is measured from the start of speech recognition to protocol prediction completion and feedback generation.
We first implement an Android application to stream the multimodal data from the AR smart glasses to the cognitive assistant pipeline for further processing. %The methodology behind the multimodal data streaming is explained in the next section. 
 The first stage of our pipeline is the speech recognition thread that continuously ingests an audio stream as input, transcribing 4 seconds of buffered audio at a time. These transcriptions are subsequently passed to the next stage for protocol prediction.

% Then, we employ the generated transcript for protocol inference. During the startup of our system, we initialize the protocol prediction model with a warm-up inference to allocate GPU resources efficiently. 
Then, the protocol prediction thread produces a ranked list of candidate protocols based on confidence scores. The protocol with the highest confidence is sent to a priority queue for immediate delivery to the responder's AR smart glasses. Next, the protocol output is combined with real-time video from the AR smart glasses for intervention recognition. Only protocols with high confidence are processed, reducing unnecessary video frame processing and increasing system efficiency. Recognized interventions are then sent to the feedback queue.
Each stage of the pipeline will be described next by highlighting, where appropriate, what was necessary to make the solution real-time and deployable on edge.

\begin{figure}[t!]
    \centering
    \includegraphics[width=\columnwidth]{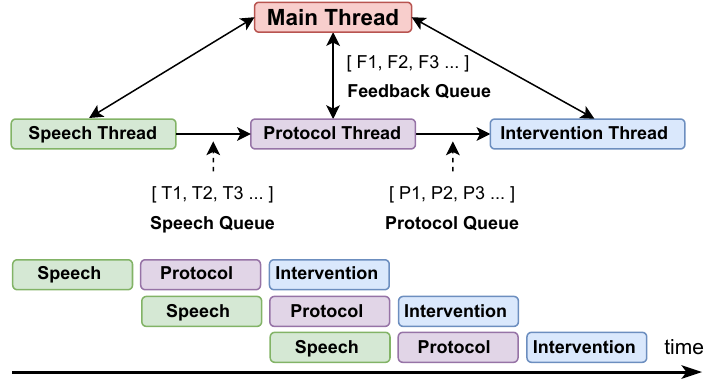} % Replace 'example-image' with your image file name and provide the correct path.
    \vspace{-1em}
    \caption{Parallel Processing at the Edge to Reduce Latency and Provide Timely Continuous Feedback}
    \label{fig:rt_pipeline}
    \vspace{-0.5em}
\end{figure}

\subsection{Multimodal Wireless Data Streaming} 
Our cognitive assistant system functions by processing continuous streams of multimodal data in real-time. In emergency situations, EMS responders will wear AR smart glasses (e.g., Vuzix M4000), which continuously transmit audio and video data to the central device of the cognitive assistant pipeline. Once the protocol is successfully predicted, the cognitive assistant relays feedback to the AR display on smart glasses. 

We have created an Android application for the AR smart glasses, responsible for three key tasks: continuous streaming of audio from the microphone to capture responders' speech, video capture from the built-in camera, and displaying feedback from the cognitive assistant on the AR display over wireless networks (Wi-Fi). Our primary goal was to optimize low-latency transmission without affecting performance. To achieve this, we designed and implemented the app using multithreading (see Figure \ref{fig:rt_pipeline}). For video and audio streams, we use the User Datagram Protocol (UDP)~\cite{postel1980user} for its lightweight, minimal packet overhead transmission. For feedback, we use the Transmission Control Protocol (TCP) for its reliable delivery, crucial for just-in-time feedback to responders. We have designed an intuitive user interface that ensures minimal interference with the responder's focus during emergencies by delivering feedback in a non-disruptive manner.

\subsection{EMS-Whisper for Speech Recognition}
To develop a speech recognition model for the EMS domain, we fine-tune the English-only tiny (39M parameters) and base (74M parameters) sizes of the Whisper family of models. We choose to build off the Whisper architecture because it has been found to be robust to noisy environments, diverse ranges of accents, and has strong out-of-the-box performance on numerous datasets~\cite{radford2023robust}. To make the two models more accurate for the EMS domain, we curate a dataset of conversational EMS speech for domain adaptation.

\begin{figure}[t!]
    \centering
    \includegraphics[width=\columnwidth]{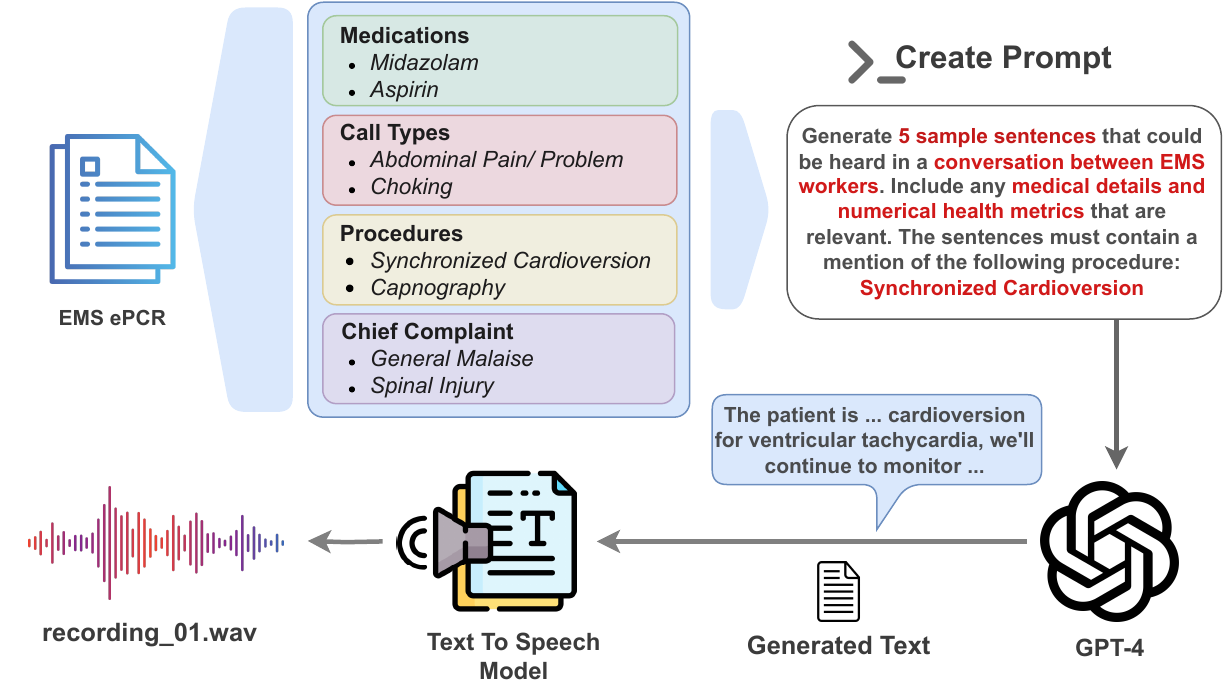} % Replace 'example-image' with your image file name and provide the correct path.
    \vspace{-1em}
    \caption{Generating Synthetic EMS Audio to Address Data Scarcity}
    \label{fig:synthetic_pipeline}
    \vspace{-1em}
\end{figure}

\shortsection{Dataset}
Our dataset to fine-tune the two Whisper models combines speech data from four different sources:

\shortsection{1. Snippets from Simulated Emergency Scenarios from EMS Responders} We consulted with several local EMS agencies and asked responders at these agencies to simulate treating patients with a variety of common patient conditions such as heart attacks, seizures, and breathing difficulties. We then recorded the simulations of enacted scenarios and manually chunked them into shorter audio segments.

\shortsection{2. Read-aloud Snippets From EMSContExt Dataset} We read aloud short segments of reports from the publicly available EMSContExt dataset~\cite{emscontext}. These texts contained vitals, medications, and symptoms commonly encountered in EMS scenes that are present in conversational EMS dialogue.

\shortsection{3. EMSAssist Training Dataset} We incorporated the publicly released EMSAssist training set, which contains numerous medical terms and phrases~\cite{jin2023emsassist}. The EMSAssist training set does not have the issue of long, incoherent sequences of medical keywords that we observe in the test set.

\shortsection{4. Synthetic Data Generated using GPT-4 and ElevenLabs Audio Generator} To train our speech recognition models on a diverse range of EMS terminology, we synthetically generate EMS conversational text using the GPT-4 Large Language Model~\cite{openai2023gpt} and synthetic audio of this text with the ElevenLabs Synthetic Voice Generator API. 

We first generate natural language prompts for GPT-4 utilizing information extracted from over 35,000 anonymized emergency Patient Care (ePCR) reports from a local urban EMS agency~\cite{kim2021information}. This extracted information contains 703 unique medications, call types, procedures, and symptoms. For each unique value, we create a prompt to ask GPT-4 to generate conversational EMS text that includes that value. We then feed this generated text to the ElevenLabs synthetic audio API and generated human-like audio with voices corresponding to a diverse range of genders, accents, and speaking rates. This synthetic audio contains rare, specialized terms that do not occur frequently, but are vital for a holistic understanding of emergency scenarios. The synthetic data generation pipeline is depicted in Figure \ref{fig:synthetic_pipeline}.

% such as: \textit{Generate 5 sample sentence that could be heard in a conversation between EMS workers. Include any medical details and numerical health metrics that are relevant. The sentences must contain a mention of the following procedure: Synchronized Cardioversion}. A sample text generation from GPT-4 is: \textit{The patient is in stable condition after synchronized cardioversion for ventricular tachycardia, we'll continue to monitor their heart rhythm closely.} 

Table \ref{tab:audiodataset} describes our train/validation/test splits and the quantity of data we obtain from each source. We publicly release the audio data from sources 2 and 4. 
% \usepackage{booktabs}
% \usepackage{graphicx}

\begin{table}
\caption{EMS Audio Dataset Curated from Different Sources}
\centering % Center the table within the column
\begin{tabular}{@{}lccccc@{}}
\toprule
\multirow{2}{*}{Split} & \multicolumn{4}{c}{Source (Duration in Minutes)} & \multirow{2}{*}{Total}  \\ 
 
 & \multicolumn{1}{c}{1} & \multicolumn{1}{c}{2} & \multicolumn{1}{c}{3} & \multicolumn{1}{c}{4} \\ \midrule
Training set & 20.8 & 0.0 & 80.4 & 41.2 & 142.4 \\
Validation set & 0.0 & 53.0 & 0.0 & \multicolumn{1}{c}{17.8} & 70.8 \\
Test set & 21.1 & 0.0 & 0.0 & 0.0 & 21.1 \\ \midrule
Total & 41.9 & 53.0 & 80.4 & 59.0 & 234.3 \\ \bottomrule
\end{tabular}
\label{tab:audiodataset}
\vspace{-1em}
\end{table}

\shortsection{Fine-tuning}
We used the curated conversational EMS dataset to fine-tune the tiny and base versions of Whisper for the EMS domain. We perform fine-tuning of Whisper with HuggingFace's Transformer's library and Trainer API~\cite{wolf2019huggingface}. We performed hyperparameter tuning using our validation set. Our final tiny model was trained for 1 epoch with a dropout rate of 0.1. Our final base model was trained for 2 epochs with a dropout of 0. Both models were trained using a batch size of 8 and a learning rate of $1 \times 10^{-5}$ with the Adam optimizer. 

We additionally train variants of the tiny and base models without synthetic data to assess whether the addition of the synthetic data improves performance. We designate Whisper fine-tuned on the full dataset with the suffix \textbf{fine-tuned} and models fine-tuned without synthetic data with the suffix \textbf{wo-synthetic}. Using our full training set, we also fine-tune the Conformer model pretrained on 1,000 hours of the LibriSpeech dataset to compare our performance with SOTA~\cite{gulati2020conformer,panayotov2015librispeech}. 

\shortsection{Edge Deployment}
To reduce computational strain on the edge device, we convert the fine-tuned Whisper models to the GGML format~\cite{ggml} with whisper.cpp~\cite{whipsercpp}, an open-source library that provides a zero-dependency high-performance C++ inference API for the Whisper models. We further optimize inference of the Encoder portion of the Whisper model using NVIDIA's cuBLAS library, which allows for partial utilization of the GPU of our edge device during inference time.

\shortsection{Streaming Setup}
We adapt whisper.cpp's open-souce stream code to ingest an audio stream from an audio device.  We process the audio stream at a sampling rate of 16,000Hz. A buffer accumulates chunks of 1024 audio samples until four seconds worth of samples (16000 x 4 = 48000) are collected. The buffer is then fed to the Whisper model for inference. This accumulation and transcription of audio data proceeds continuously until the cognitive assistant is terminated. We utilize multiprocessing to handle the audio buffering task to maximize efficiency and parallelism and ensure low latency.

\subsection{EMS-TinyBERT for Protocol Selection}

We propose the EMS-TinyBERT which can be deployed on edge devices for accurate real-time protocol selection. Our model %design mainly follows \cite{ge2023dkec} which includes 
consists of three parts as shown in Figure~\ref{fig:ems_protocol_selection}: a text encoder to encode EMS transcripts (from ePCR or input audio) into text features; a graph neural network to fuse domain knowledge with text features; a group-wise training strategy to deal with scarcity of training samples in rare classes.

\shortsection{Text Encoder} We adopt TinyClinicalBERT~\cite{rohanian2023lightweight}, a BERT model pretrained on medical corpus, as the text encoder. After removing the punctuations and stopwords in the EMS transcripts $t$, we use TinyClinicalBERT to encode the transcript and take the last hidden states of TinyClinicalBERT as text features $\mathbf{E}_{t} \in \mathbb{R}^{\mbox{seq}\times\delta}$, where $\mbox{seq}$ is the length of text and $\delta$ is the dimension of hidden states in TinyClinicalBERT. %as follows,

%\begin{equation}
%    \mathbf{E}_{t} = \mbox{TinyClinicalBERT}(t)
%\end{equation}
%\vspace{-1em}

\begin{figure}[t!]
    \centering
    \includegraphics[width=\columnwidth]{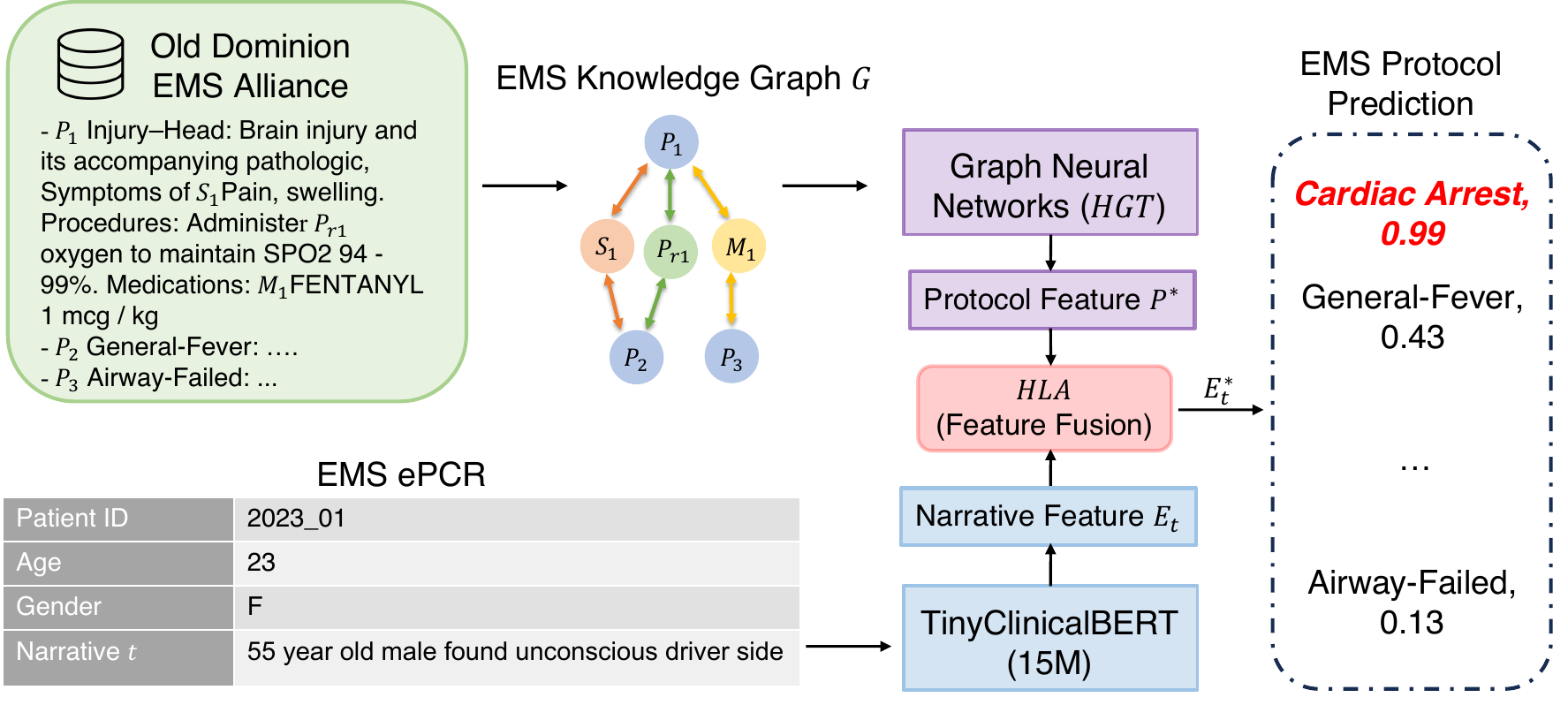} % Replace 'example-image' with your image file name and provide the correct path.
    \caption{EMS-TinyBERT: A Tiny Model that Fuses EMS Domain Knowledge (Old Dominion EMS Alliance) to Improve Protocol Prediction Performance}
    \vspace{-1em}
    \label{fig:ems_protocol_selection}
    \vspace{-1em}
\end{figure}

\shortsection{Domain Knowledge Fusion}
To utilize external medical domain knowledge for EMS protocol prediction, we follow the methods in~\cite{ge2023dkec} to construct a heterogeneous graph $G$ and utilize a feature fusion module to integrate domain knowledge into text features for classification.
% then a feature fusion module is utilized to combine text and domain knowledge together for classification.

ODEMSA is utilized as medical guidelines in our works, which contain detailed descriptions of different medical conditions, diverse relations between medical entities like signs and symptoms of specific conditions, and the interventions that medical professionals should perform to provide consistent patient care. Specifically, we manually extract the symptoms, medications, and procedures for every EMS protocol and then construct a graph $G = \left(N, E\right)$, with $N$ as the set of nodes (protocols, medications, symptoms, procedures) and $E$ as the set of edges. We use GatorTron~\cite{yang2022gatortron} to generate initial node embeddings based on the descriptions or names of the nodes in the protocol guidelines. Then we adopt a 1 layer heterogeneous graph transformer (HGT)~\cite{hu2020heterogeneous} to capture the relations between different protocol nodes in graph $G$. The output of $\mbox{HGT}$ are updated node embeddings ($\mathbf{P}^{\star} = \mbox{HGT}(G)$), from which we only use the updated protocol features for feature fusion.

Then a heterogeneous label-wise attention~\cite{ge2023dkec} (HLA) module takes both protocol features $\mathbf{P}^{\star}$ and text feature $E_t$ as input and fuses the features by having the protocols assign different weights for each token in the text representation ($\mathbf{E}^{\star}_{t} = \mbox{HLA}(\mathbf{P}^{\star}, \mathbf{E}_t)$).
%\begin{align}
%    \mathbf{P}^{\star} &= \mbox{HGT}(G)\\
%    \mathbf{E}^{\star}_{t} &= \mbox{HLA}(\mathbf{P}^{\star}, \mathbf{E}_t)
%\end{align}
Finally, the fused features $\mathbf{E}^{\star}_{t}$ are flattened and fed into a linear layer for classification.

\shortsection{Group-wise Training}
In the EMS domain, there are often different protocols for the treatment of the same condition for patients with different pre-existing conditions (e.g., pediatric vs. adult patients). While the signs and symptoms for such protocols are similar to those of adults, the guidelines on specific interventions that are appropriate for children are different. Therefore, we group the highly similar protocols(e.g.: \textit{Medical-seizure (adult protocol 3-12)} and \textit{Medical-seizure (pediatric protocol 9-12)}) during the model training phase for coarse-grained classification. During inference, we used pre-defined rules (e.g.: age $\leq$ 18 indicates a pediatric case) to do fine-grained classification. We use the sigmoid function to normalize the model's final output to be between 0 and 1. A binary cross entropy loss is used to measure the distance between prediction and ground truth. The final output from the model is the EMS protocol predictions and their corresponding confidence scores (see Figure~\ref{fig:ems_protocol_selection}).

\shortsection{Edge Deployment} We implement two strategies to shorten the inference time on the edge device. First, we used a lightweight 15M parameter pretrained model, TinyClinicalBERT, as our text encoder. Secondly, to reduce the time of constructing an EMS knowledge graph, we store all the initial node embeddings in the memory so that the model does not need to repetitively generate initial embeddings during inference. Moreover, during the system startup, we initialize the protocol prediction model with a warm-up inference to allocate GPU resources efficiently to minimize runtime overheads. %For real-time EMS protocol prediction in the edge device, we accumulate the chunked transcripts from the output of EMS-Whisper to feed into EMS-TinyBERT.

\shortsection{Streaming Setup} EMS-TinyBERT is capable of receiving and processing text input in chunks during the inference phase, enabling real-time prediction based on streaming textual transcripts from audio data. These textual chunks are progressively accumulated over the course of the EMS responder interactions to achieve a holistic understanding of the incident context for more accurate prediction. We utilize this ability to support end-to-end evaluations of our system.

\begin{figure}
    \centering
    \includegraphics[width=\columnwidth]{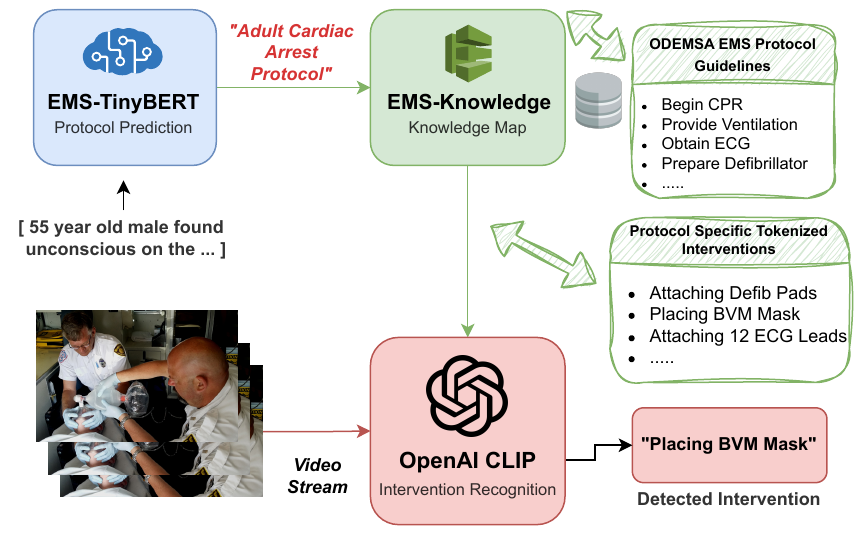} % Replace 'example-image' with your image file name and provide the correct path.
    \vspace{-2em}
    \caption{Integrating EMS Protocol Knowledge to Improve Intervention Recognition Module Performance}
    \label{fig:emsvisionarch}
    \vspace{-0.75em}
\end{figure}

\subsection{EMS-Vision for Intervention Recognition}

% Our approach to intervention recognition Figure \ref{fig:emsvisionarch} includes 3 critical components. First, the speech data from the responder is processed by the speech recognition module followed by the protocol prediction module. The protocol predictions, in turn, serve as a contextual basis. Utilizing a knowledge agent established upon protocol guidelines, we subsequently provide a subset of interventions associated with the protocol to the vision module. The vision module consists of a Zero-shot image classification model proposed by \cite{radford2021learning} which takes in image frames from the video along with the candidate labels for interventions. The comprehensiveness of EMS protocols encompasses an extensive array of interventions and presents a challenge for efficient for effective inference by a zero-shot classification model. Therefore, the identification of a pertinent subset of interventions, succinctly described in natural language, is of paramount significance to fully exploit the potential of the CLIP model. 

Our intervention recognition approach (Figure \ref{fig:emsvisionarch}) consists of three subcomponents. The protocol predictions from EMS-TinyBERT serve as a contextual basis, and a knowledge agent established upon protocol guidelines provides a subset of interventions associated with the protocol to the vision module. The vision module utilizes a zero-shot image classification model, CLIP proposed by~\cite{radford2021learning}, which takes in video frames and candidate labels for interventions. To fully exploit the potential of the CLIP model, identifying a pertinent and concise subset of interventions in natural language is critical, considering the comprehensiveness of EMS protocols that encompass extensive interventions.

For our preliminary implementation, we consider two protocols \textit{Medical - chest pain - cardiac suspected (protocol 2 - 1)} and \textit{Medical - respiratory distress/asthma/copd/croup/reactive airway (respiratory distress)} defined in ODEMSA.
%to collect data of simulated scenarios. 
%We create a knowledge agent as depicted in \ref{fig:emsvisionarch} which takes the predicted protocol as input to provide associated interventions with the protocol as input to the image classification model. Moreover, 
For every 4 seconds of multimodal audio/video data, the output with the highest confidence from the protocol prediction module is fed to a knowledge agent as depicted in \ref{fig:emsvisionarch}. The knowledge agent will provide the associated interventions for the protocol as input to the CLIP model. The CLIP model will then generate a predicted intervention output for every frame in the window. The windows where the protocol prediction confidence is low will not have any intervention predictions. The knowledge agent will provide interventions described in natural language to provide CLIP with an image-text pair that encompasses a stronger semantic relationship while minimizing the similarity with other non-matching pairs.

%An additional objective of our research is to deploy our cognitive assistant on edge computing devices. 
\shortsection{Edge Deployment} The initial configuration of the CLIP model demands computational resources and dependencies that are not available on our edge device, the NVIDIA Jetson Nano. Hence, we utilize a lightweight implementation of CLIP written in C/C++ to support CPU-Only inference designed for resource-constrained devices~\cite{clipcpp}. Moreover, we apply 8-bit integer quantization~\cite{nagel2021white} which effectively transforms the trained model to represent weights and activations of layers in the neural net with 8-bit integers over high-precision values such as 32-bit floating point numbers. This significantly reduces the model size which essentially leads to lower memory requirements and faster inference times.

\section{Evaluation}
We perform a comprehensive evaluation of the performance and latency of the standalone modules and the end-to-end pipeline compared to EMSAssist~\cite{jin2023emsassist} models, including EMSConformer for speech recognition and EMSMobileBERT for protocol prediction, using the following datasets and metrics.

\subsection{Evaluation Datasets}

\textbf{Electronic Patient Care Report (ePCR) Dataset} is a collection of 4,417 anonymized pre-hospital ePCRs obtained from a local urban ambulance agency. Each ePCR contains first responders’ textual descriptions of the patient's situation, interventions performed, patient’s medical history, and EMS protocols used.
% We used protocol labels present in the ePCR as ground truth when evaluating performance of our protocol prediction model.

\textbf{EMS Audio Dataset} is comprised of two subsets.
    The first subset is used for standalone evaluation of speech recognition models to assess the effect of fine-tuning on model performance. This subset consists of short pre-chunked segments of audio, as is standard in general speech recognition datasets. 
    The second subset is used to evaluate the speech recognition and protocol selection models in the context of our end-to-end pipeline. This subset contains longer post-incident narratives, along with transcripts and protocol labels for each narrative. 
% For both these subsets, we use ground truth transcripts of the audio as reference text when evaluating our speech models.

\textbf{EMS Video Dataset} is a collection of 40 videos (equating to approximately 17000 images) of two distinct intervention actions: `Defibrillation' using AED and `Supporting Airway' using a Bag Valve Mask (BVM). Defibrillation is relevant for cardiac-related emergencies, and BVM is relevant to respiratory distress emergencies. This dataset is used for the standalone evaluation of the intervention recognition module.
%Labels for this video frames in this dataset were set based on the intervention that was performed in the associated scenario.

\textbf{EMS Multimodal Dataset} consists of eight video and audio scenarios simulating a realistic EMS incident from start to end. This dataset is used for an integrated evaluation of speech recognition and intervention recognition. To collect this dataset, we first developed eight realistic EMS incident scenarios for cardiac-related and respiratory distress emergencies by referencing several example EMS training scenarios from Alabama Public Health~\cite{AlabamaProtocol} and the EMS Online~\cite{CBTProtocol} websites, combined with a team member's background knowledge from their time as an EMT. We then acted out the scenarios in a simulated setting with a dummy patient and recorded synchronized audio and video using the CognitiveEMS pipeline. %(Because this dataset did not have a representative distribution of protocol labels, we did not use it for the evaluation of the protocol prediction module).
Due to insufficient incident details in the conversational part of this dataset, the protocol prediction module could not function as intended on the small chunks of transcribed audio, resulting in low performance. So we decided to not use this dataset for the end-to-end evaluation of the protocol prediction module.
% Labels for the video frames in this dataset were set based on the intervention that was performed in the associated scenario. We used the oiginal transcripts of the scenarios as reference text when evaluating speech recognition performance.

\subsection{Evaluation Metrics}

\shortsection{Speech Recognition}
We use the standard metrics of Word Error Rate (WER) and Character Error Rate (CER) to assess the performance of our speech recognition module. When calculating metrics for the Whisper models, we first normalize the predictions using the model's tokenizer, which standardizes punctuation and special characters. When calculating metrics for SOTA's EMSConformer, we first preprocess the ground truth transcripts remove all punctuation except for periods, and convert numbers to their phonetic representations because the EMSConformer has a limited vocabulary.

We also report latency measurements (in milliseconds) for the different speech recognition models when they are run as part of the end-to-end pipeline. These measurements correspond to the time it takes for the model to transcribe one chunk of audio, which we have chosen to be 4 seconds long.

\shortsection{EMS Protocol Prediction}
We used multiple metrics for the evaluation of the protocol selection module to have a fair and complete comparison to SOTA. Threshold-based metrics like Micro F1($miF$) and Macro F1($maF$) are used by setting the threshold as 0.5. Ranking based metrics like top-k accuracy($Acc@K$) which do not require a specific threshold are also reported. Since the average number of labels per instance in the ePCR dataset is 1.2, we set $K=1$ for top-k accuracy($Acc@K$), but also report $Acc@3$. We describe these metrics and their drawbacks below:

\begin{itemize}
    \item $miF$ is heavily influenced by frequent protocols and thus can be used to evaluate the overall performance.
    \item $maF$ weighs the F1 score achieved on each protocol equally and is used to evaluate the performance for the rare protocols.
    \item $Acc@K$ computes the number of times where the correct label is among the top k  predicted labels, which are ranked by confidence scores.
\end{itemize}

\shortsection{Intervention Recognition} We use the following metrics for the evaluation of intervention recognition module. These metrics are used 
to assess the impact of protocol knowledge input on the intervention recognition and the end-to-end real-time performance. 
\begin{itemize}
    \item  \textit{Accuracy} refers to the ratio of video frames classified as the correct intervention to all video frames associated with the intervention. Each video frame is labeled by a human with the intervention taking place at that moment, and this is used to identify whether the intervention recognition outputs agree with the ground truth.
    \item   \textit{Latency} refers to the duration between the retrieval of a video frame by the vision module and the generation of recognized intervention.
\end{itemize}

\shortsection{End-to-End Evaluation} We evaluate the performance and latencies of each module in the pipeline when operated in an end-to-end setting. We use the same metrics defined for each module with the addition of latency measurement. Moreover, we report two overall latencies for protocol feedback generation and intervention recognition in an end-to-end execution.

\subsection{Experimental Platforms} We implement our cognitive assistant system on two platforms for performance comparison and end-to-end evaluation.

\shortsection{Edge Device}
We utilize a Jetson Nano, an Edge AI apparatus manufactured by NVIDIA. The Jetson Nano provides an ARM A57 Quad Core processor with a clock speed of 1.43Ghz, RAM of 4GB, and a 128-Core Maxwell GPU running on Ubuntu 18.04 with CUDA 10.2.  Our evaluation only involves the execution of the inference phase of the models, with no involvement of the edge device in the training process.

\shortsection{Server}
We utilize a server configured with a 13th Generation Intel(R) Core(TM) i9-13900KF processor, operating at a clock speed of 3.0 GHz and featuring a total of 32 CPU cores. This server is complemented by 32 gigabytes of system RAM and accommodates an NVIDIA RTX 4080 GPU, which possesses 9728 CUDA cores with 16 gigabytes of dedicated GPU memory running on Ubuntu 20.04.

\subsection{EMS-Whisper Evaluation} We perform two evaluations of our ASR models. First, we evaluate the models' standalone performance with short pre-chunked segments of audio, as is common in standard evaluations of ASR systems. We perform this evaluation on the server using the test split of our EMS Audio dataset.

% \todo[inline]{Not sure if this paragraph should go here or in the End-to-End System}
We also assess the performance of the speech recognition models when integrated into the end-to-end pipeline and processing a stream of audio in parallel with the other modules. For this second evaluation, we use the full-length conversational post-incident EMS reports from subset (2) of the aforementioned EMS Audio Dataset and the recordings from the EMS Multimodal dataset. We also calculate the WER and CER of the combined transcription output of the entire report (which is the finalized transcripts of each chunk concatenated together). We also calculate the average latency to process every 4-second chunk of audio. We do not report results for Whisper base results on the Jetson edge device because the model exhausted the computational resources of the device when run in parallel with other modules of the pipeline. We also only report metrics for the EMSConformer on the server because we encountered compatibility issues when implementing the model on the NVIDIA Jetson.

\subsection{EMS-TinyBERT Evaluation}
We use the scikit-multilearn~\cite{szymanski2017scikit} to split the ePCR dataset by the ratio of 70:30 for training and testing and further perform 3-fold cross-validation. The evaluation results are based on the average performance of the three test sets to alleviate the bias introduced by data splitting. There are in total 43 EMS protocol labels in our dataset. 

We used GatorTron~\cite{yang2022gatortron} to generate the initial embeddings for all the nodes in the knowledge graph. The Heterogeneous Graph Transformer(HGT)~\cite{hu2020heterogeneous} was adopted as the GNN model for domain knowledge fusion. We set the number of layers in HGT as 1, and the hidden dimension in HGT is set as 256. We use Adam optimizer for training with batch sizes ranging from 4 to 32 and learning rates ranging from 1e-6 to 1e-3 for hyper-parameter tuning. To avoid over-fitting, we use regularization with the weight decay of 1e-5, a dropout rate of 0.3, and early stopping if the validation loss keeps increasing more than three times. We implemented the EMS-TinyBERT based on the Huggingface~\cite{wolf2019huggingface} and the PyTorch~\cite{paszke2019pytorch}. The model was trained with one NVIDIA GPU RTX3090.

\subsection{EMS-Vision Evaluation}

Using the intervention videos dataset, we evaluate the performance on the server with and without protocol knowledge to assess the impact of integrating the knowledge when using zero short models. Moreover, we only evaluate intervention recognition using our fine-tuned speech models to avoid redundant results. Then, we employ the end-to-end  Scenarios dataset featuring simulated interactions between EMS responders and a patient encountering cardiac-related and respiratory distress emergency scenarios, which were collected to assess the performance of our intervention recognition system under more realistic conditions. This evaluation is part of the end-to-end system evaluation explained in the next section.

\subsection{End-to-End System}

We deploy the cognitive assistant on two platforms, as previously described. The edge-deployed version, however, features fundamental differences in the implementation of certain components, such as speech recognition and action recognition, aimed at optimizing computational resource utilization while ensuring acceptable response times. For evaluation purposes, we simulate audio streaming from AR smart glasses by creating a separate process that reads an audio file and writes it to a virtual microphone interface through PulseAudio~\cite{pulseaudio}, a networked low-latency sound server for Linux and POSIX systems. In the case of video streaming, we load the video data and transmit frames at their original rate to mimic real-time video feed, accomplished using OpenCV to ensure the intervention recognition module's proper functioning~\cite{bradski2000opencv}.

% We utilize a dataset consisting of ten audio recordings obtained from an EMS agency, including authentic post-incident narratives and reports, as well as simulated EMS conversations with patients, along with self-collected scenarios mentioned earlier. The ground truth data includes the speech transcripts and established protocols for reference.
We utilized ten recordings acquired from an EMS agency in the Audio Dataset to conduct an end-to-end assessment of the Speech Recognition module, followed by the Protocol Prediction module. However, since we lack access to the videos associated with these recordings, we are unable to evaluate the Intervention Recognition module. As an alternative, we use the Multimodal Dataset, which consists of 8 self-collected scenarios, to evaluate the end-to-end performance from speech recognition to intervention recognition.

Furthermore, for a more comparative analysis of our cognitive assistant pipeline, we implement SOTA \cite{jin2023emsassist} speech recognition module (EMSConformer) and protocol prediction module (EMSMobileBERT) on our server. However, due to certain incompatibilities in software dependencies, we could not implement and evaluate their models on our edge device. %However, we encountered certain incompatibilities in software dependencies, which hindered us from implementing their speech recognition module on the edge device. 
In order to ensure a fair comparison, we trained and fine-tuned their models on our own datasets. Also, for performance comparison at the edge, we use the best results of their model performance on our server to our results at the edge. This is assuming that their model has the same or better performance on the server than the edge as reported in \cite{jin2023emsassist}. Performance metrics and latencies for each stage/module in our pipeline are reported in Tables \ref{tab:e2eperf} and \ref{tab:e2elatency}. Additionally, we provide an overall latency measurement for protocol prediction feedback and intervention recognition.

\section{Results}

% For our reference
% \begin{enumerate}
%     \item A fine-tuned Tiny speech recognition model that is optimized for Edge devices achieving real-time performance.
%   \item Speech recognition outperforms state of the art significantly on EMS conversations with WER of \textbf{0.162} vs 0.441
%   \item Our Tiny speech model significantly outperforms SOTA tiny model by WER \textbf{0.186} vs 0.458 while maintaining a comparable latency \colorbox{yellow}{INSERT NUMBERS}
%   \item Speech Recognition improves with Synthetic data used during fine-tuning. WER \textbf{0.162} vs 0.170 
%   \item A tiny protocol selection model outperforms SOTA with higher accuracy (micro f1: 0.756 vs 0.216) and fast inference time.
%   \item Intervention recognition shows a significant improvement in accuracy of 30\% when accurate protocol knowledge is provided to the vision model.
%   \item Intervention recognition achieves similar accuracies on edge vs server (0.727 vs 0.743). However, due to resource constraints on edge, it achieved a higher latency than expected.
%   \item Our overall end-to-end latency for speech recognition and protocol prediction is considerably lower than the SOTA solution (\textbf{3.7s} vs 4.2s)
  
% \end{enumerate}

\subsection{EMS Speech Recognition}
\shortsection{Standard Speech Recognition Evaluation}
We first discuss the results of the traditional evaluation of the speech recognition models on the test split of our curated EMS Speech dataset. As shown in Tabel \ref{tab:standard_ASR}, we observe that \textbf{Whisper models fine-tuned with both human and synthetic data achieve the best performance}. Our tiny-fine-tuned model achieves a WER of 0.152 while the original tiny model achieves a WER of 0.156. Our base-fine-tuned model achieves a WER of 0.122 while the original base achieves a WER of 0.133. Both of our fine-tuned Whisper models achieve substantially lower error rates compared to SOTA's fine-tuned base and lite models (with WER of 0.355 and 0.436, respectively). These results demonstrate that fine-tuning the pretrained Whisper models using human and synthetically-generated audio can improve transcription accuracy for conversational EMS speech. We also find that the \textbf{tiny-fine-tuned model offers the best tradeoff} between model size and WER for edge deployment.

% \usepackage{booktabs}
\begin{table}[t!]
\caption{Standard ASR Evaluation on EMS Audio Dataset}
\centering
\begin{adjustbox}{height=1.8cm}
    
\begin{tabular}{@{}lrr@{}}
\toprule
Model & \multicolumn{1}{l}{WER} & \multicolumn{1}{l}{CER} \\ \midrule
tiny & 0.156 & 0.093 \\
tiny-fine-tuned-wo-synthetic & 0.158 & 0.094 \\
\textbf{tiny-fine-tuned} (39M) & \textbf{0.152} & \textbf{0.091} \\
\midrule
base & 0.133 & 0.079 \\
base-fine-tuned-wo-synthetic & 0.131 & 0.079 \\
\textbf{base-fine-tuned} (74M) & \textbf{0.122} & \textbf{0.077} \\
\midrule
EMSConformer-tflite (10M) & 0.436 & 0.250 \\
EMSConformer-base (10M) & 0.355 & 0.203 \\  \bottomrule
\end{tabular}
\end{adjustbox}
\label{tab:standard_ASR}
\vspace{-2em}
\end{table}

% \shortsection{Streaming End-to-End Evaluation}

\subsection{EMS Protocol Prediction}
We compare our protocol prediction model, EMS-TinyBERT, with SOTA models including EMSMobileBERT, TinyClinicalBERT in terms of performance and model size. Results show that EMS-TinyBERT achieves a good trade-off between memory usage and performance. Specifically, \textbf{it outperforms SOTA methods by more than 10\% in performance on ePCR dataset, while only introducing 6\%} overhead in number of parameters (15.9M vs. 15M).

As shown in Table~\ref{tab:RAA results}, EMS-TinyBERT has the best performance over all the baselines. For frequently used EMS protocols, EMS-TinyBERT has the highest $miF$ 0.756, which outperforms EMSMobileBERT (0.216) and TinyClinicalBERT (0.673), showing that EMS-TinyBERT can provide correct EMS protocols for first responders when encountering common conditions. For rarely used protocols, EMS-TinyBERT has the best $maF$ score (0.499) over all other models, indicating its good ability to provide EMS protocol recommendations for rare conditions. Using ranking based evaluation metrics, EMS-TinyBERT achieves the best performance among all models. $Acc@1$ in EMS-TinyBERT indicates that 76.5\% top1 recommendation is correct. Furthermore, $Acc@3$ of EMS-TinyBERT is 90.4\%, indicating there is 90.4\% chance that ground truth EMS protocols are in the top-3 predictions of our model. The comparison also shows that EMSMobileBERT does not perform well on real-world ePCR data and can not be applied to EMS conversational scenarios. 

We also validate the necessity of each module in the design of EMS-TinyBERT model.  Table~\ref{tab:RAA results} shows that both Group-wise training and knowledge fusion help to improve the model's ability in protocol prediction. It is notable that the knowledge fusion module will introduce an extra 0.9M parameters to the raw model because the graph neural network is used to incorporate EMS domain knowledge. However, compared with baseline TinyClinical(15M), it is acceptable to have a 6\% increase in the number of parameters to achieve more than 10\% performance improvement. Besides, EMS-TinyBERT(15.9M) has better performance and less overhead than EMSMobileBERT(25M) on edge devices.

\begin{table}[]
    \caption{Protocol Selection comparison to State-of-the-art on ePCR dataset, EMS-TinyBERT (15.9M) includes both Knowledge Fusion and Group-wise Training on ePCR Dataset.}
    \centering
    \resizebox{\columnwidth}{!}{
    \begin{tabular}{lcccc}
        \toprule
        \multicolumn{1}{c}{Model (Size)} & $miF$ & $maF$ & $Acc@1$ & $Acc@3$ \\
         \cmidrule{1-5}
          TinyClinicalBERT (15M) & 0.673 & 0.171 & 0.679 & 0.807 \\
         EMSMobileBERT (25M) & 0.216 & 0.034 & 0.226 & 0.439 \\
         \cmidrule{1-5}
         TinyClinicalBERT-Group (15M) & 0.709 & 0.277 & 0.696 & 0.836 \\
         TinyClinicalBERT-Knowledge Fusion (15.9M) & 0.743 & 0.395 & 0.744 & 0.898 \\
         \textbf{EMS-TinyBERT} (15.9M) & \textbf{0.756} & \textbf{0.499} & \textbf{0.765} & \textbf{0.904} \\
        \bottomrule
    \end{tabular}}
    \label{tab:RAA results}
    \vspace{-1em}
\end{table}

\subsection{EMS Intervention Recognition}
To assess the effect of using multimodal data, we compare the results of intervention recognition obtained with and without the knowledge of protocol. Table \ref{tab:intervention} shows two examples of recognizing \textit{Attaching Defibrillator} in \textit{Cardiac Arrest} protocol and \textit{Placing Oxugen Mask} in \textit{Respiratory Distress} protocol. %by incorporating the protocol knowledge (based on ground truth protocol labels) as input to intervention recognition. Then, we perform the intervention recognition on the same data without the protocol knowledge (only video data). 
When incorporating the protocol knowledge (from ground truth labels), the CLIP model is provided with a subset of natural language tokens describing the interventions that are relevant to the target protocol. But without such knowledge, the CLIP model will consider the full set of possible EMS interventions.  
We see that by \textbf{incorporating the protocol knowledge, the accuracy of the intervention recognition improves significantly (\textbf{28\% (0.79 vs. 0.51)}-42\% (0.64 vs. 0.22))}. As indicated in Table \ref{tab:intervention}, without the protocol knowledge, the CLIP model faces difficulties in identifying the most suitable intervention that describes the image, resulting in a higher number of false positives as there are too many candidate interventions to consider.

\begin{table}[t!]
\caption{Performance of Intervention Recognition with and without Prior Knowledge of the Protocol on EMS Video Dataset
}
\resizebox{\columnwidth}{!}{%
\begin{tabular}{@{}lrlr@{}}
\toprule
\multicolumn{2}{c}{With Protocol Knowledge} &
  \multicolumn{2}{c}{Without Protocol Knowledge} \\ \midrule
  Intervention &
  Accuracy   &
  Intervention &
  \multicolumn{1}{l}{Accuracy} \\ \midrule
  
\textbf{Attaching Defibrillator} & \multicolumn{1}{r}{\textbf{0.79}} &
Attaching Defibrillator & 0.51 \\
Inserting IV to arm & \multicolumn{1}{r}{0.11} & Attaching nebulizer & 0.16 \\
 Inserting IV to leg  & \multicolumn{1}{r}{0.07} & Defibrillator & 0.09  \\
Defibrillator & \multicolumn{1}{r}{0.02}  & Inserting IV to arm   & 0.07  \\
\midrule
\textbf{Placing Oxygen mask on face} & \multicolumn{1}{r}{\textbf{0.64}} & Placing Oxygen mask on face & 0.22  \\
Attaching nebulizer & \multicolumn{1}{r}{0.21} & Inserting IV to arm & 0.19 \\
Inserting airway adjunct   & 0.07   & Attaching two Defib pads on chest  & 0.15 \\
Administering albuterol   & 0.03   & Inserting IV to leg  & 0.15 \\
 \bottomrule
\end{tabular}
}
\label{tab:intervention}
\vspace{-1em}
\end{table}

\begin{table*}[t!]
\caption{Average End-to-End Performance of CognitiveEMS Pipeline on Edge Devices vs. Servers Compared to SOTA}
\vspace{-0.5em}
% \caption{fine-tuning improves }
\resizebox{\textwidth}{!}{%
\begin{threeparttable}
\begin{tabular}{@{}llcccccccccccc@{}}
\toprule

\multirow{4}{*}{Speech Model} &\multirow{4}{*}{Protocol Model}  & \multicolumn{10}{c}{EMS Audio Dataset} & \multicolumn{2}{c}{EMS Multimodal Dataset} \\ \cmidrule(lr){3-12} \cmidrule(lr){13-14}

 &  & \multicolumn{4}{c}{Speech Recognition} & \multicolumn{6}{c}{Protocol Prediction} & \multicolumn{2}{c}{Intervention Recognition} \\ %\cmidrule(r){3-14} 

 &  & \multicolumn{2}{c}{Server} & \multicolumn{2}{c}{Edge} & \multicolumn{3}{c}{Server} & \multicolumn{3}{c}{Edge} & Server & Edge \\  \cmidrule(r){3-6}  \cmidrule(lr){7-12}  \cmidrule(l){13-14}
 &  & \multicolumn{1}{c}{WER} & \multicolumn{1}{c}{CER} & \multicolumn{1}{c}{WER} & \multicolumn{1}{c}{CER} & $miF$ & $Acc@1$ & $Acc@3$ & $miF$ & $Acc@1$ & $Acc@3$ & Accuracy & Accuracy \\ \cmidrule{1-14}
 &  & \multicolumn{1}{l}{} & \multicolumn{1}{l}{} & \multicolumn{1}{l}{} & \multicolumn{1}{l}{} & \multicolumn{1}{l}{} & \multicolumn{1}{l}{} & \multicolumn{1}{l}{} & \multicolumn{1}{l}{} & \multicolumn{1}{l}{} & \multicolumn{1}{l}{} & \multicolumn{1}{l}{} & \multicolumn{1}{l}{} \\ 
tiny.en & EMS-TinyBERT & 0.276 & 0.176 & 0.301 & 0.201 & 0.550 & 0.550 & 0.700 & \textbf{0.650} & \textbf{0.650} & \textbf{0.800} & - & - \\

tiny-wo-synthetic & EMS-TinyBERT & 0.271 & 0.175 & 0.290 & 0.195 & 0.600 & 0.600 & \textbf{0.800} & 0.600 & 0.600 & 0.700 & - & - \\

\textbf{tiny-fine-tuned} & EMS-TinyBERT & \textbf{0.266} & \textbf{0.162} & \textbf{0.290} & \textbf{0.192} & \textbf{0.600} & \textbf{0.600} & 0.700 & 0.600 & 0.600 & 0.750 & \textbf{0.743} & \textbf{0.727} \\
\midrule
base.en & EMS-TinyBERT & 0.281 & 0.186 & - & - & \textbf{0.650} & \textbf{0.650} & 0.800 & - & - & - & - & - \\

base-wo-synthetic & EMS-TinyBERT & 0.261 & 0.175 & - & - & 0.500 & 0.500 & 0.800 & - & - & - & - & - \\

\textbf{base-fine-tuned} & EMS-TinyBERT & \textbf{0.225} & \textbf{0.152} & - & - & 0.600 & 0.600 & \textbf{0.850} & - & - & - & \textbf{0.775} & - \\

\midrule
EMSConformer-tflite* & EMSMobileBERT & 0.618 & 0.440 & \multicolumn{1}{c}{-} & \multicolumn{1}{c}{-} & 0.071 & 0.100 & 0.200 & - & - & - & - & -
\\
EMSConformer-tf-base* & EMSMobileBERT & 0.591 & 0.429 & \multicolumn{1}{c}{-} & \multicolumn{1}{c}{-} & 0.020 & 0.056 & 0.111 & - & - & - & - & -

\\ \bottomrule
\end{tabular}%
\begin{tablenotes}
\item[*] Based on our implementation of EMSConformer and EMSMobileBERT \cite{jin2023emsassist} models using our data and server. Not evaluated on our edge device due to incompatibilities.
\end{tablenotes}
\end{threeparttable}
}
\label{tab:e2eperf}
\end{table*}

\begin{table*}[t!]
\vspace{-1em}
\caption{Average End-to-End Latency of CognitiveEMS Pipeline on Edge Devices vs. Servers Compared to SOTA}
\vspace{-0.5em}
\resizebox{\textwidth}{!}{%
\begin{threeparttable}
\begin{tabular}{@{}llcccccccccc@{}}

\toprule
\multirow{3}{*}{Speech Model} & \multirow{3}{*}{Protocol Model} & \multicolumn{10}{c}{Average Latency (s)} \\ \cmidrule{3-12} 
 &  & \multicolumn{2}{c}{Speech Recognition} & \multicolumn{2}{c}{Protocol Prediction} & \multicolumn{2}{c}{Intervention Recognition} & \multicolumn{2}{c}{Protocol Feedback} & \multicolumn{2}{c}{Intervention Feedback} \\

 &  & Server & Edge & Server & Edge & Server & Edge & Server & Edge & Server & Edge \\
  \cmidrule{3-12}  
 \cmidrule(r){1-2}
 &  & \multicolumn{1}{c}{} & \multicolumn{1}{c}{} & \multicolumn{1}{l}{} & \multicolumn{1}{l}{} & \multicolumn{1}{l}{} & \multicolumn{1}{l}{} & \multicolumn{1}{l}{} & \multicolumn{1}{l}{} & \multicolumn{1}{l}{} & \multicolumn{1}{l}{} \\

tiny-fine-tuned & EMS-TinyBERT & 0.16 & \textbf{3.65} & 0.01 & \textbf{0.14} & 0.02 & 4.46 & 0.18 & \textbf{3.78} & 0.19 & 8.24 \\

base-fine-tuned & EMS-TinyBERT & 0.30 & - & \textbf{0.01} & 0.18 & 0.02 & - & 0.31 & - & 0.32 & - \\
\midrule
EMSConformer-tf-lite & EMSMobileBERT & \textbf{0.10} & 3.60* & 0.01 & 0.60* & - & - & \textbf{0.11} & 4.20* & - & - \\ 
EMSConformer-base & EMSMobileBERT & 0.19 & - & 0.02 & - & - & - & 0.21 & - & - & - \\

\bottomrule
\end{tabular}%
\begin{tablenotes}
\item[*] As reported in \cite{jin2023emsassist} on an Essential Phone PH-1 and not evaluated on our edge device due to incompatibilities.
\end{tablenotes}
\end{threeparttable}
}
\label{tab:e2elatency}
\vspace{-0.5em}
\end{table*}

\subsection{End-to-End System}

We evaluate the end-to-end performance and latency of our cognitive assistant pipeline, including a comparative analysis of our proposed speech and protocol prediction models and the previous SOTA. In Tables \ref{tab:e2eperf} and \ref{tab:e2elatency}, we provide performance metrics and latencies for each stage of the pipeline.

\shortsection{Speech Recognition Performance and Latency}
In the end-to-end pipeline, we observe that our \textbf{tiny-fine-tuned model achieves the lowest WER of 0.290 on the Jetson edge device} and also outperforms SOTA's fine-tuned Conformer model running on the server (which had a WER of 0.591). Our \textbf{base-fine-tuned model achieves the lowest WER on the server} across all models with a WER of 0.225 and achieves a substantially lower WER rate than SOTA's base model (0.591).

We also observe that our \textbf{fine-tuned-tiny model achieves a 3.65 second latency on our edge device} when processing a 4-second chunk of audio, which is comparable to the latency of SOTA's EMS-Conformer (3.60 seconds).

Our speech results demonstrate that our fine-tuned Whisper models outperform SOTA and the original Whisper models on conversational EMS audio in a real-time streaming setup. We also demonstrate that our tiny-fine-tuned model runs effectively on our edge device in conjunction with other modules of our cognitive assistant pipeline.
\ref{tab:e2eperf}.

\shortsection{Protocol Prediction Performance and Latency}
Our results demonstrate a significant improvement in protocol prediction performance compared to the SOTA (EMSMobileBERT) both on the server and the edge device. On the server, \textbf{the best $miF$ of our EMS-TinyBERT is 0.650 compared with 0.071 on EMSAssist, the best $Acc@1$ of EMS-TinyBERT is 0.650 compared with 0.100 on EMSAssist, and the best $Acc@3$ of EMS-TinyBERT is 0.850 compared with 0.200 on EMSAssit. The latency on the server of our EMS-TinyBERT is 0.01s compared with 0.02s on EMSAssist}. Compared with  EMSAssist, our EMS-TinyBERT takes less inference time and achieves better protocol prediction performance. There are several reasons for the superiority of our protocol prediction model. First, compared with EMSAssist, EMS-TinyBERT has fewer parameters (15M) than EMSMobileBERT (25M), which shortens the inference time on both the server and the edge device. However, EMS-TinyBERT achieves better EMS protocol prediction because of utilizing a medical pretrained model TinyClinicalBERT as the backbone and incorporating EMS domain knowledge for decision making.

\shortsection{Intervention Recognition Performance and Latency}
% The results indicate that on the Edge device, the accuracy is slightly lower compared to the server, with 0.727 vs. 0.743 accuracy, respectively. This difference in accuracy can be attributed to the changes we applied to the base model to accommodate the resource constraints on the Edge device. Due to the limited resources available, we observe a significantly higher latency on the Edge device (4456.61 ms) compared to the server (16.32 ms). This is primarily due to the inability to use the base CLIP model that utilizes GPU resources to accelerate the inference. Instead, the entire inference process on the Edge device is performed on the CPU, which has very low compute capacity.
Notably, intervention recognition is able to \textbf{achieve comparable accuracy on both edge and server, 0.727 vs. 0.743, respectively}. This difference in accuracy can be attributed to the changes we applied to the base model to accommodate the resource constraints on the edge device.  Due to the limited resources available, we observe a significantly higher latency on the edge device (4.46s) compared to the server (0.02s). This is primarily due to the inability to use the base CLIP model that utilizes GPU resources to accelerate the inference. Instead, the entire inference process on the edge device is performed on the CPU, which has low compute capacity and remains a limitation for this edge device.
Moreover, on the server, the base fine-tuned speech model has been shown to indirectly improve intervention recognition accuracy \textbf{0.775} vs. \textbf{0.743} through more accurate transcriptions, which are utilized by the downstream protocol prediction model to output more accurate protocol predictions.

\shortsection{End-to-End Latency} Our results show that the \textbf{end-to-end latency for protocol feedback outperforms SOTA with a considerably lower value of \textbf{3.78s} vs. 4.20s on the edge}, while meeting the pre-defined deadline of \textbf{4s} for the real-time system. While speech recognition latencies are comparable to SOTA (\textbf{3.65s} vs. 3.60s), the latency for protocol prediction is significantly reduced (\textbf{0.14s} vs. 0.60s), thereby contributing to an overall reduction in latency.

\section{Conclusion}
% <talk about future work here too>
This paper presents a novel real-time cognitive assistant deployed on edge that utilizes multimodal data to continuously support EMS responders during emergencies. With a fine-tuned speech recognition model for EMS conversations and a protocol prediction model built using a tiny language model incorporating EMS domain knowledge and the novel addition of intervention recognition using multimodal data, our cognitive assistant demonstrates significant performance improvement compared to SOTA.
Despite using real-world data in our evaluations, the real-world deployment of the proposed system in EMS training or actual incidents requires further improvement, refinements, and evaluation in consultation with the first responders. Future work will focus on improving the real-time accuracy and performance and validating the practical utility and user-friendliness of our cognitive assistant system through real-life incident simulations and feedback integration in collaboration with our partner EMS agencies.

\section{Acknowledgement}
This work was supported in part by the award 70NANB21H029 from the U.S. Department of Commerce, National Institute of Standards and Technology (NIST).
\bibliographystyle{IEEEtran}
\bibliography{root}

\end{document}